\documentclass{article}

\usepackage[final]{cpal_2026}

\usepackage{times}
\usepackage{soul}
\usepackage{url}
\usepackage[hidelinks]{hyperref}
\usepackage[utf8]{inputenc}
\usepackage[small]{caption}
\usepackage{graphicx}
\usepackage{amsmath}
\usepackage{amsthm}
\usepackage{booktabs}
\usepackage{algorithm}
\usepackage{algorithmic}

\usepackage{bbm}
\usepackage[export]{adjustbox}
\usepackage{lipsum}
\usepackage{color}
\usepackage{wrapfig}
\usepackage{booktabs}
\usepackage{multirow,mathtools} 
\usepackage{threeparttable}
\usepackage{amsfonts}
\usepackage{wrapfig}
\usepackage{times}
\usepackage{multirow}
\usepackage{url}
\usepackage{xcolor}

\usepackage{colortbl}
\usepackage{wrapfig}
\usepackage{tikz}
\usepackage{pifont}
\usepackage{booktabs, makecell, tabularx}
\usepackage{rotating}
\newcommand{\gr}{\rowcolor[gray]{.95}}

\title{(\texttt{PASS}) Visual \underline{P}rompt Loc\underline{a}tes Good \underline{S}tructure \underline{S}parsity \\ through a Recurrent HyperNetwork}

\author{%
  Tianjin Huang\textsuperscript{1,2}\thanks{These authors contributed equally.}, ~Yong Tao\textsuperscript{1}\footnotemark[1],~Meng Fang\textsuperscript{3},~Li Shen\textsuperscript{4},~Fan Liu\textsuperscript{5},~Yulong Pei\textsuperscript{2} \\ ~Mykola Pechenizkiy\textsuperscript{2},~Tianlong Chen\textsuperscript{6}\\
  \textsuperscript{1}University of Exeter,~\textsuperscript{2}Eindhoven University of Technology,~\textsuperscript{3}University of liverpool\\~\textsuperscript{4}Sun Yat-sen University,~\textsuperscript{5}Hohai University,~\textsuperscript{6}UNC-Chapel Hill\\
  \texttt{\{t.huang2, yt438\}@exeter.ac.uk, Meng.Fang@liverpool.ac.uk, mathshenli@gmail.com, fanliu@hhu.edu.cn, \{y.pei.1, M.pechenizkiy\}@tue.nl, tianlong@mit.edu}
}

\begin{document}

\maketitle

\begin{abstract}
Large-scale neural networks have demonstrated remarkable performance in different domains like vision and language processing, although at the cost of massive computation resources. As illustrated by compression literature, structural model pruning is a prominent algorithm to encourage model efficiency, thanks to its acceleration-friendly sparsity patterns. One of the key questions of structural pruning is how to estimate the channel significance. In parallel, work on data-centric AI has shown that prompting-based techniques enable impressive generalization of large language models across diverse downstream tasks. In this paper, we investigate a charming possibility - \textit{leveraging visual prompts to capture the channel importance and derive high-quality structural sparsity}. To this end, we propose a novel algorithmic framework, namely \texttt{PASS}. It is a tailored hyper-network to take both visual prompts and network weight statistics as input, and output layer-wise channel sparsity in a recurrent manner. Such designs consider the intrinsic channel dependency between layers. Comprehensive experiments across multiple network architectures and six datasets demonstrate the superiority of \texttt{PASS} in locating good structural sparsity. For example, at the same FLOPs level, \texttt{PASS} subnetworks achieve $1\%\sim 3\%$ better accuracy on Food101 dataset; or with a similar performance of $80\%$ accuracy, \texttt{PASS} subnetworks obtain $0.35\times$ more speedup than the baselines.
\end{abstract}

\section{Introduction}
Recently, large-scale neural networks, particularly in the field of vision and language modeling, have received upsurging interest due to the promising performance for both natural language~\cite{brown2020language,chiang2023vicuna} and vision tasks~\cite{dehghani2023scaling,bai2023qwen}. While these models have delivered remarkable performance,  their colossal model size, coupled with their vast memory and computational requirements, pose significant obstacles to model deployment. To solve this daunting challenge, model compression techniques have re-gained numerous attention~\cite{dettmers2022llm,xiao2023smoothquant} and knowledge distillation can be further adopted on top of them to recover optimal performance~\cite{huang2023large,sun2019patient}. Among them, model pruning is 
a well-established method known for its capacity to reduce model size without compromising performance and structural model pruning has garnered significant interest due to its ability to systematically eliminate superfluous structural components, such as entire neurons, channels, or filters, rather than individual weights, making it more hardware-friendly~\cite{li2017pruning,fang2023depgraph}. 

In the context of structural pruning for \textit{vision models}, the paramount task is the estimation of the importance of each structure component, such as channel or filters.  It is a fundamental challenge since it requires dissecting the neural network behavior and a precise evaluation of the relevance of individual structural sub-modules. Previous methodologies~\cite{liu2017learning,fang2023depgraph} have either employed heuristics or developed learning pipelines to derive scores, achieving notable performance. Recently, the prevailingness of natural language prompts  has facilitated an emerging wisdom that the success of AI is deeply rooted in the quality and specificity of data that is originally created by human~\cite{zha2023data}. Techniques such as in-context learning~\cite{chen2022improving}  and prompting~\cite{razdaibiedina2023progressive} have been developed to create meticulously designed prompts or input templates to escalate the output
quality of LLMs. These strategies bolster the capabilities of LLMs and consistently achieve notable success across diverse downstream tasks. This offers a brand new angle for addressing the intricacies of structural pruning on importance estimation of vision models:
\textit{How can we leverage the potentials within the input space to facilitate the dissection of the relevance of each individual structural component across layers, thereby enhancing structural sparsity?} 

One straightforward approach is directly editing input through visual prompt to enhance the performance of compressed vision models~\cite{xu2023compress}. The performance upper bound of this approach largely hinges on the quality of the sparse model achieved by pruning, given that prompt learning is applied post-pruning.  Moreover, when pruning is employed to address the intricate relevance between structural components across layers, the potential advantages of using visual prompts are not taken into consideration. 

Therefore, we posit that probing judicious input editing is imperative for structural pruning to examine the importance of structural components in vision models. The \textbf{crux of our research} lies in embracing an innovative \underline{\textbf{data-centric}} viewpoint towards structural pruning. Instead of designing or learning prompts on top of compressed models, we develop a novel end-to-end framework for channel pruning, which identifies and retains the most crucial channels across models by incorporating visual prompts, referred to as \textbf{\texttt{PASS}}. 

\begin{figure*}[bt]
\centering
\includegraphics[width=0.8\textwidth]{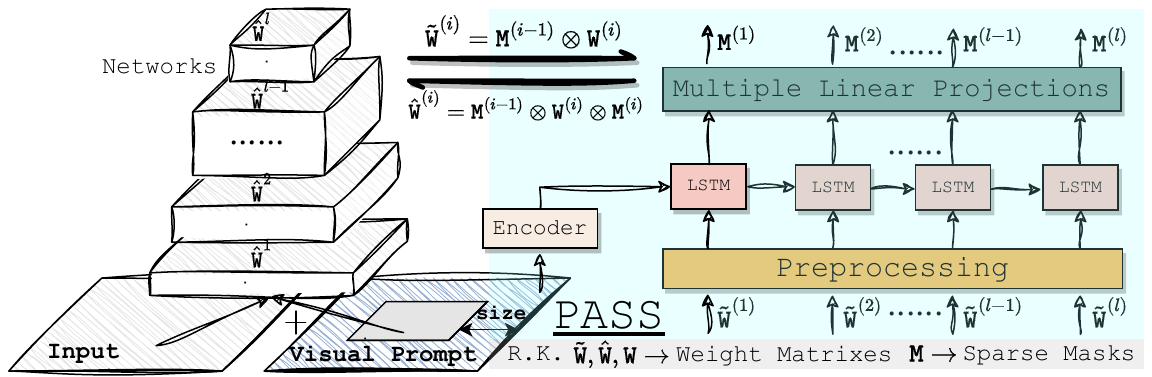}
\caption{The overall framework of \texttt{PASS}. (\textit{Left}) Our pruning target is a convolutional neural network (CNN) that takes images and visual prompts as input. (\textit{Right}) The \texttt{PASS} hyper-network integrates the information from visual prompts and layer-wise weight statistics, then determines the significant structural topologies in a recurrent fashion.}
\label{fig:framework}
\end{figure*}

Moreover, the complexities associated with inherent channel dependencies render the generation of sparse channel masks a challenging task. Due to this reason, many previous arts of pruning design delicate pruning metrics to recognize sparse subnetworks with smooth gradient flow~\cite{wang2020picking}. To better handle the channel dependencies across layers during channel pruning, we propose to learn sparse masks using a \textbf{recurrent mechanism}. Specifically, the learned sparse mask for the recent layer largely depends on the mask from the previous layer in an efficient recurrent manner, and all the masks are learned by incorporating the extra information provided by visual prompts. The \texttt{PASS} framework is shown in Figure~\ref{fig:framework}. Our contributions are summarized as follows:

\begin{itemize}
    \item We probe and comprehend the role of the input editing in the context of channel pruning, and confirming the imperative to integrate visual prompts for crucial channel discovery. 
    \item  To handle the complex dependence caused by channel elimination across layers, we further develop a recurrent mechanism to efficiently learn layer-wise sparse masks by taking both the sparse masks from previous layers and visual prompts into consideration. Anchored by these innovations, we propose \texttt{PASS}, a pioneering framework dedicated to proficient channel pruning in convolution neural networks from a data-centric perspective. 
    \item  Through comprehensive evaluations across six datasets containing \{CIFAR-10, CIFAR-100, Tiny-ImageNet, Food101, DTD, StanfordCars\} and four architectures including \{ResNet-18, ResNet-34, ResNet-50, VGG\}, our results consistently demonstrate \texttt{PASS}'s significant potential in enhancing both the performance of the resultant sparse models and computational efficiency.

    \item More interestingly, our empirical studies reveal that the sparse channel masks and the hypernetwork produced by \texttt{PASS}  exhibit superior transferability, proving beneficial for a range of subsequent tasks.
\end{itemize}

\section{Related Work}
\textbf{Structural Network Pruning.}
Structural pruning achieves network compression through entirely eliminating certain superfluous components from the dense network. In general, structural pruning follows three steps: (i) pre-training a large, dense model; (ii) pruning the unimportant channels based on criteria, and (iii) finetuning the pruned model to recover optimal performance.
The primary contribution of various pruning approaches is located in the second step: proposing proper pruning metrics to identify the importance of channels. Some commonly-used pruning metric includes but not limited to weight norm~\cite{li2016pruning}, Taylor expansion~\cite{molchanov2016pruning}, feature-maps reconstruction error~\cite{he2018amc} KL-divergence~\cite{luo2020neural}, greedy forward selection with largest loss reduction~\cite{ye2020good}, feature-maps discriminant information \cite{hou2020feature}.


\noindent\textbf{Prompting.}
In the realm of natural language processing, prompting has been acknowledged as an effective strategy to adapt pre-trained models to specific tasks \cite{liu2023pre}. The power of this technique was highlighted by GPT-3's successful generalization in transfer learning tasks using carefully curated text prompts \cite{brown2020language}. Researchers have focused on refining text prompting methods \cite{shin2020autoprompt} and developed a technique known as Prompt Tuning. This approach involves using prompts as task-specific continuous vectors optimized during fine-tuning \cite{li2021prefix}, offering comparable performance to full fine-tuning with a significant reduction in parameter storage and optimization. Prompt tuning's application in the visual domain has seen significant advancement recently. Pioneered by  \cite{bahng2022exploring}, who introduced prompt parameters to input images, the concept was expanded by \cite{chen2023understanding} to envelop input images with prompt parameters.  \cite{jia2022visual} took this further, proposing visual prompt tuning for Vision Transformer models. Subsequently, \cite{liu2023explicit} designed a prompt adapter to enhance these prompts. Concurrently, \cite{zang2022unified} integrated visual and text prompts in vision-language models, boosting downstream performance.

\noindent\textbf{Hypernetwork.}
Hypernetworks represent a specialized form of network architecture, specifically designed to generate the weights of another Deep Neural Network (DNN). This design provides a meta-learning approach that enables dynamic weight generation and adaptability, which is crucial in scenarios where flexibility and learning efficiency are paramount. Initial iterations of hypernetworks, as proposed by ~\cite{zhang2018graph}, were configured to generate the weights for an entire target DNN. While this approach is favorable for smaller and less complex networks, it constrains the efficacy of hypernetworks when applied to larger and more intricate ones. To address this limitation, subsequent advancements in hypernetworks have been introduced, such as the component-wise generation of weights ~\cite{zhao2020meta} and chunk-wise generation of weights ~\cite{chauhan2023dynamic}. Diverging from the initial goal of hypernetworks, our work employs them to fuse visual prompts and model information for generating sparse channel masks.

\section{\texttt{PASS}: Visual \underline{P}rompt Loc\underline{a}tes Good \underline{S}tructure \underline{S}parsity}

\textbf{Notations.} Let us consider a CNN with $l$ layers, and each layer $i$ contains its corresponding weight tensor $\mathtt{W}^{(i)} \in \mathbb{R}^{\mathrm{C}_{\mathrm{O}}^{i} \times \mathrm{C}_{\mathrm{I}}^{i} \times \mathrm{K}^{i} \times \mathrm{K}^{i}}$, where \{$\mathrm{C}_{\mathrm{O}}^{i}$, $\mathrm{C}_{\mathrm{I}}^{i}$, and $\mathrm{K}^{i}$\} are the number of output/input channels and convolutional kernel size, respectively. The entire parameter space for the network is defined as $\mathtt{W}=\{\mathtt{W}^{(i)}\}|_{i=1}^{l}$. Similarly, a layer-wise binary mask is represented by $\mathtt{M}^{(i)}$, where ``$0$"/``$1$" indicates removing/maintaining the associated channel. $\mathtt{V}$ denotes our visual prompts. $(\boldsymbol{x},y ) \in \mathcal{D} $ denotes the data of a target task.

\noindent\textbf{Rationale.} In the realm of structural pruning for deep neural networks, one of the key challenges is how to derive channel-wise importance scores for each layer. Conventional mechanisms estimate the channel significance either in a global or layer-wise manner~\cite{he2017channel,fang2023depgraph}, neglecting the sequential dependency between adjacency layers. Meanwhile, the majority of prevalent pruning methods are designed in a \textit{model-centric} fashion~\cite{fang2023depgraph}. In contrast, an ideal solution to infer the high-quality sparse mask for one neural network layer $i$ should satisfy several conditions as follows:
\begin{itemize} 
    \item [\ding{172}] \textit{$\mathtt{M}^{(i)}$ should be dependent to $\mathtt{M}^{(i-1)}$}. The sequential dependency between layers should be explicitly considered. It plays an essential role in encouraging gradient flow throughout the model~\cite{wang2020picking,pham2022understanding}, by preserving structural ``pathways".
    \item [\ding{173}] \textit{$\mathtt{M}^{(i)}$ should be dependent to $\mathtt{W}^{(i)}$}. The statistics of network weights are commonly appreciated as powerful features for estimating channel importance~\cite{liu2017learning,li2016pruning}.
    \item [\ding{174}] \textit{$\mathtt{M}^{(i)}$ should be dependent to $\mathtt{V}$}. Motivated by the \textit{data-centric} advances in NLP, such prompting can contribute to the dissecting and understanding of model behaviors~\cite{razdaibiedina2023progressive,chen2022knowprompt}.
\end{itemize}
Therefore, it can be expressed as $\mathtt{M}^{(i)} = f(\mathtt{M}^{(i-1)}, \mathtt{W}^{(i)}, \mathtt{V})$, where the generation of a channel mask for layer $i$ depends on the weights in the current layer, the previous layers' mask, and visual prompts.

\subsection{Innovative Data-Model Co-designs through A Recurrent Hypernetwork}
To meet the aforementioned requirements, \texttt{PASS} is proposed as illustrated in Figure~\ref{fig:framework}, which enables the data-model co-design pruning via a recurrent hyper-network. Details are presented below.

\noindent\textbf{Modeling the Layer Sequential Dependency.} The recurrent hyper-network in \texttt{PASS} adopts a Long Short-Term Memory (LSTM) backbone since it is particularly suitable for capturing sequential dependency. It enables an ``auto-regressive" way to infer the structural sparse mask. To be specific, the LSTM mainly utilizes the previous layer's mask $\mathtt{M}^{(i-1)}$, the current layer's weights $\mathtt{W}^{(i)}$, and a visual prompt $\mathtt{V}$ as follows:
\begin{equation}
\mathtt{M}^{(i)} = \texttt{LSTM}_{\theta}( \mathtt{\widetilde{W}}^{i}, g_{\omega}(\mathtt{V}))  \; \mathtt{M}^{(0)}=\texttt{LSTM}(\mathtt{W}^{(i)}, g_{\omega}(\mathtt{V})),
\label{equ:hypernetwork}
\end{equation}
where $\widetilde{\mathtt{W}}^{(i)}=\mathtt{M}^{(i-1)}\otimes \mathtt{W}^{(i)}$, the visual prompt $\mathtt{V}$ provides an initial hidden state for the LSTM hyper-network, $\theta$ is the parameters of the LSTM model, and $g_{\omega}(\mathtt{V})$ is the extra encoder to map the visual prompt into an embedding space. The channel-wise sparse masks ($\mathtt{M}^{(i)}$) generated from the hyper-network are utilized to prune the weights of each layer as expressed by $\widehat{\mathtt{W}}^{(i)}=\mathtt{M}^{(i-1)}\otimes \mathtt{W}^{(i)}\otimes \mathtt{M}^{(i)}$. $\mathtt{M}^{(i-1)}\otimes \mathtt{W}^{(i)}$ represents the pruning of in-channels while $\mathtt{W}^{(i)}\otimes \mathtt{M}^{(i)}$ denotes the pruning of out-channels.

\noindent\textbf{Visual Prompt Encoder.} An encoder is used to extract representations from the raw visual prompt $\mathtt{V}$. $g_{\omega}(\mathtt{V})$ denotes a three-layer convolution network and $\omega$ are the parameters for the CNN $g_{\omega}(\cdot)$. The dimension of extracted representations equals the dimension of the hidden state of the LSTM model. A learnable embedding will serve as the initial hidden state for the LSTM model.

\noindent\textbf{Preprocessing the Weight.} The in-channel pruned weights $\widetilde{\mathtt{W}}^{(i)}$ is a $4$D matrix. In order to take this weight information, it is \underline{first} transformed into a vector of length equal to the number of out-channels by averaging the weights over the $\mathrm{C}_{\mathrm{I}}^{i} \times \mathrm{K}^{i} \times \mathrm{K}^{i} $ dimensions. \underline{Then}, these vectors are padded by zero elements to unify their length.

\noindent\textbf{Converting Embedding to Channel-wise Sparse Mask.} Generating layer-wise channel masks from the LSTM module presents two challenges: ($1$) it outputs embeddings of a uniform length, whereas the number of channels differs at each layer; ($2$) producing differentiable channel masks directly from this module is infeasible.  To tackle these issues, \texttt{PASS} adopts a two-step approach: \ding{172} An independent linear layer is employed to map the learned embeddings onto channel-wise important scores corresponding to each layer.  \ding{173} During the forward pass in training, the binary channel mask $\mathtt{M}$ is produced by setting the $(1-s)\times100\%$ elements with the highest channel-wise important scores to $1$, with the rest elements set to $0$. In the backward pass, it is optimized by leveraging the straight-through estimation method~\cite{bengio2013estimating}. Here the $s\in(0,1)$ denotes the channel sparsity of the network layer.

For achieving an optimal non-uniform layer-wise sparsity ratio, we adopt \underline{global pruning}~\cite{huang2022you} that eliminates the channels associated with the lowest score values from all layers during each optimization step. This approach is grounded in the findings of \cite{huang2022you,fang2023depgraph}, which demonstrate that layer-wise sparsity derived using this method surpasses other extensively researched sparsity ratios.

\subsection{How to Optimize the Hypernetwork in \texttt{PASS}}

\textbf{Learning \texttt{PASS}.} The procedures of learning \texttt{PASS} involves a jointly optimization of the visual prompt $\mathtt{V}$, encoder weights $\omega$, and \texttt{LSTM}'s model weights $\theta$. Formally, it can be described below:
\begin{equation}
    \min_{\theta,\omega, \mathtt{V}}  \mathcal{L}(\Phi_{\widehat{\mathtt{W}}}(\boldsymbol{x}+\mathtt{V}),y), \; \widehat{\mathtt{W}}^{(i)}=\mathtt{M}^{(i-1)}\otimes \mathtt{W}^{(i)}\otimes \mathtt{M}^{(i)},
\end{equation}
Where $\Phi_{\widehat{\mathtt{W}}}(\cdot)$ is the target CNN with weights $\widehat{\mathtt{W}}$, $\boldsymbol{x}$ and $y$ are the input image and its groundtruth label. Note that $\mathtt{M}^{(i)}$ is generated by $\texttt{LSTM}_{\theta}(\mathbf{\widetilde{\mathtt{W}}}^{i}, g_{\omega}(\mathtt{V}))$ as described in Equation~\ref{equ:hypernetwork}. The objective of this learning phase is to optimize the \texttt{PASS} model to generate layer-wise channel masks, leveraging both a visual prompt $\mathtt{V}$ and the inherent model weight statistics as guidance. After that, the obtained sparse subnetwork will be further fine-tuned on the downstream dataset.  

\noindent\textbf{Fine-tuning Sparse Subnetwork.} The procedures of subnetwork fine-tuning involve the optimization of the visual prompt $\mathtt{V}$ and model weights $\mathtt{W}$, which can be expressed by:
\begin{equation}
    \min_{\mathtt{W}, \mathtt{V}}  \mathcal{L}(\Phi_{\widehat{\mathtt{W}}}(\boldsymbol{x}+\mathtt{V}),y),
\end{equation}
where $\widehat{\mathtt{W}}=\mathtt{M}^{(i-1)}\otimes \mathtt{W}^{(i)}\otimes \mathtt{M}^{(i)}$ and the sparse channel mask $\mathtt{M}$ is fixed.

\section{Experiments}
In this section, we empirically demonstrate the effectiveness of our proposed \texttt{PASS} method against various baselines across multiple datasets and models. Additionally, we evaluate the transferability of the sparse channel masks and the hypernetwork learned by \texttt{PASS}. Further, we validate the superiority of our specific design by a series of ablations studies. 

To evaluate \texttt{PASS}, we follow the widely-used evaluation of visual prompting which is pre-trained on large datasets and evaluated on various target domains~\cite{chen2023understanding,jia2022visual}.
Specifically, this process is accomplished by two steps: ($1$) Identifying an optimal structural sparse neural network based on a pre-trained model and ($2$) Fine-tuning the structural sparse neural network on the target task.  During the training process, we utilize the Frequency-based Label Mapping \(FLM\) as presented by ~\cite{chen2023understanding} to facilitate the mapping of the logits from the pre-trained model to the logits of the target tasks. 

\subsection{Implementation Setups} \label{exp_setup}
 \textbf{Architectures and Datasets.} We evaluate \texttt{PASS} using four traditional pre-trained models: ResNet-$18$, ResNet-$34$, ResNet-$50$, and VGG-$16$ without BatchNorm$2$D and three advanced models:ResNeXt-50,ViT-B/16 and Swin-T, all pre-trained on ImageNet-1K. Our evaluation contains six target tasks: Tiny-ImageNet, CIFAR-$10/100$, DTD, StanfordCars, and Food$101$. 

\noindent\textbf{Baselines.} We select five popular structural pruning methods as our baselines: ($1$) \textit{Group-L$1$ structural pruning}~\cite{li2017pruning} reduces the network channels via $l_{1}$ regularization. ($2$) \textit{GrowReg}~\cite{wang2021neural} prunes the network channels via $l_{2}$ regularization with a growing penalty scheme.  ($3$) \textit{Slim}~\cite{liu2017learning} imposes channel sparsity by applying $l_{1}$ regularization to the scaling factors in batch normalization layers. ($4$) \textit{DepGraph}~\cite{fang2023depgraph} models the inter-layer dependency and group-coupled parameters for pruning and ($5$) \textit{ABC Pruner}~\cite{lin2020channel} performs channel pruning through automatic structure search.  

\noindent\textbf{Training and Evaluation.} We utilize off-the-shelf models from Torchvision~\footnote{\url{https://pytorch.org/vision/stable/index.html}} as the pre-trained models. During the pruning phase, we employ the SGD optimizer for the visual prompt, while the AdamW optimizer is used for the visual prompt encoder and the LSTM model for generating channel masks. Regarding the baselines, namely Group-L$1$ structural pruning, GrowReg, Slim, and DepGraph, they are trained based on this implementation~\footnote{\url{https://github.com/VainF/Torch-Pruning}} and ABC Prunner is trained based on their official public code~\footnote{\url{https://github.com/lmbxmu/ABCPruner}}. During the fine-tuning phase, all pruned models, inclusive of those from \texttt{PASS} and the aforementioned baselines, are fine-tuned with the same hyper-parameters. 
For all experiments, we report the accuracy of the downstream task during testing and the floating point operations (FLOPs) for measuring the efficiency.

\subsection{\texttt{PASS} Finds Good Structural Sparsity}
In this section, we first validate the effectiveness of \texttt{PASS} across multiple downstream tasks and various model architectures. Subsequently, we investigate the transferability of both the generated channel masks and the associated model responsible for generating them.

\begin{figure*}[tb]
\centering
\includegraphics[width=0.97\textwidth]{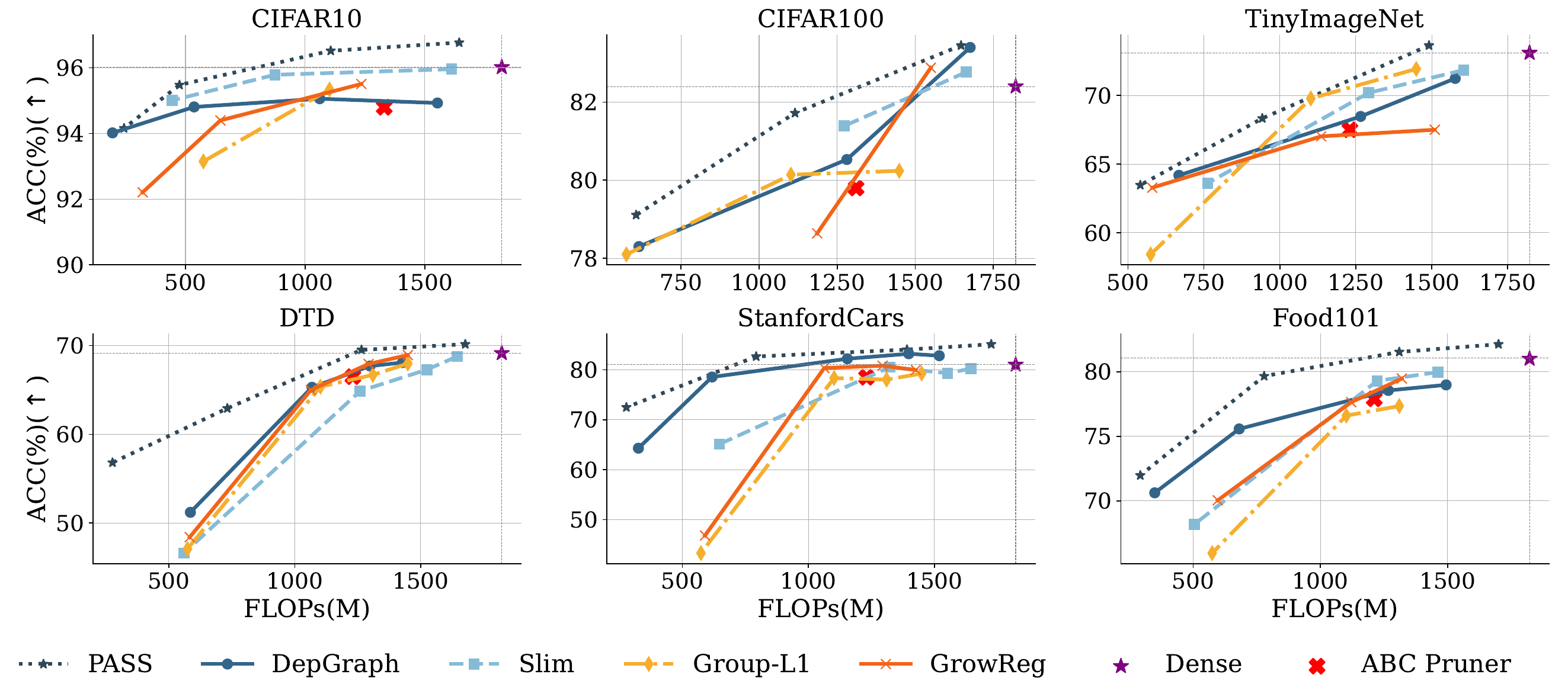}
\vskip -0.1 in
\caption{ Test accuracy of channel-pruned networks across multiple downstream tasks based on the pre-trained ResNet-18.}
\label{fig:downstreams}
\vskip -0.15 in
\end{figure*}

\noindent\textbf{Superior Performance across  Downstream Tasks}.
In Figure~\ref{fig:downstreams}, we present the test accuracy of the \texttt{PASS} method in comparison with several baseline techniques, including Group-L$1$, GrowReg, DepGraph, Slim, and ABC Prunner. The evaluation includes six downstream tasks: CIFAR-$10$, CIFAR-$100$, Tiny-ImageNet, DTD, StanfordCars, and Food$101$. The accuracies are reported against varying FLOPs to provide a comprehensive understanding of \texttt{PASS}'s efficiency and performance. 

From Figure~\ref{fig:downstreams}, several salient observations can be drawn: \ding{182}  \texttt{PASS} consistently demonstrates superior accuracy across varying FLOPs values for all six evaluated downstream tasks. \underline{On one hand}, \texttt{PASS} achieves higher accuracy under the same FLOPs. For example, it achieves $1\%\sim 3\%$ higher accuracy than baselines under $1000 \mathtt{M}$ FLOPs among all the datasets. \underline{On the other hand}, \texttt{PASS} attains higher speedup\footnote{Following~\cite{fang2023depgraph}, we report the theoretical speedup ratios and it is defined as \(\frac{\text{FLOPs}_{\texttt{PASS}} - \text{FLOPs}_{\texttt{baseline}}}{\text{FLOPs}_{\texttt{baseline}}}\)} in achieving comparable accuracy levels. For instance, to reach accuracy levels of \(96\%\), \(81\%\), and \(80\%\) on CIFAR$10$, StanfordCars, and Food$101$ respectively, the \texttt{PASS} method consistently realizes a speedup of at least \(0.35\times\) (\(900\) VS \(1400\)), outperforming the most competitive baseline. This consistent performance highlights the robustness and versatility of the \texttt{PASS} method across diverse scenarios. \ding{183}  In terms of resilience to pruning, \texttt{PASS} exhibits a more gradual reduction in accuracy as FLOPs decrease. This trend is notably more favorable when compared with the sharper declines observed in other baseline methods. \ding{184} Remarkably, at the higher FLOPs levels, \texttt{PASS} not only attains peak accuracies but also surpasses the performance metrics of the fully fine-tuned dense models. For instance, \texttt{PASS} excels the fully fine-tuned dense models with \{$1.05\%$, $0.99\%$, $1.06\%$\} on CIFRAR$100$, DTD and FOOD$101$ datasets. 

\begin{figure*}[tb]
\centering
\includegraphics[width=0.97\textwidth]{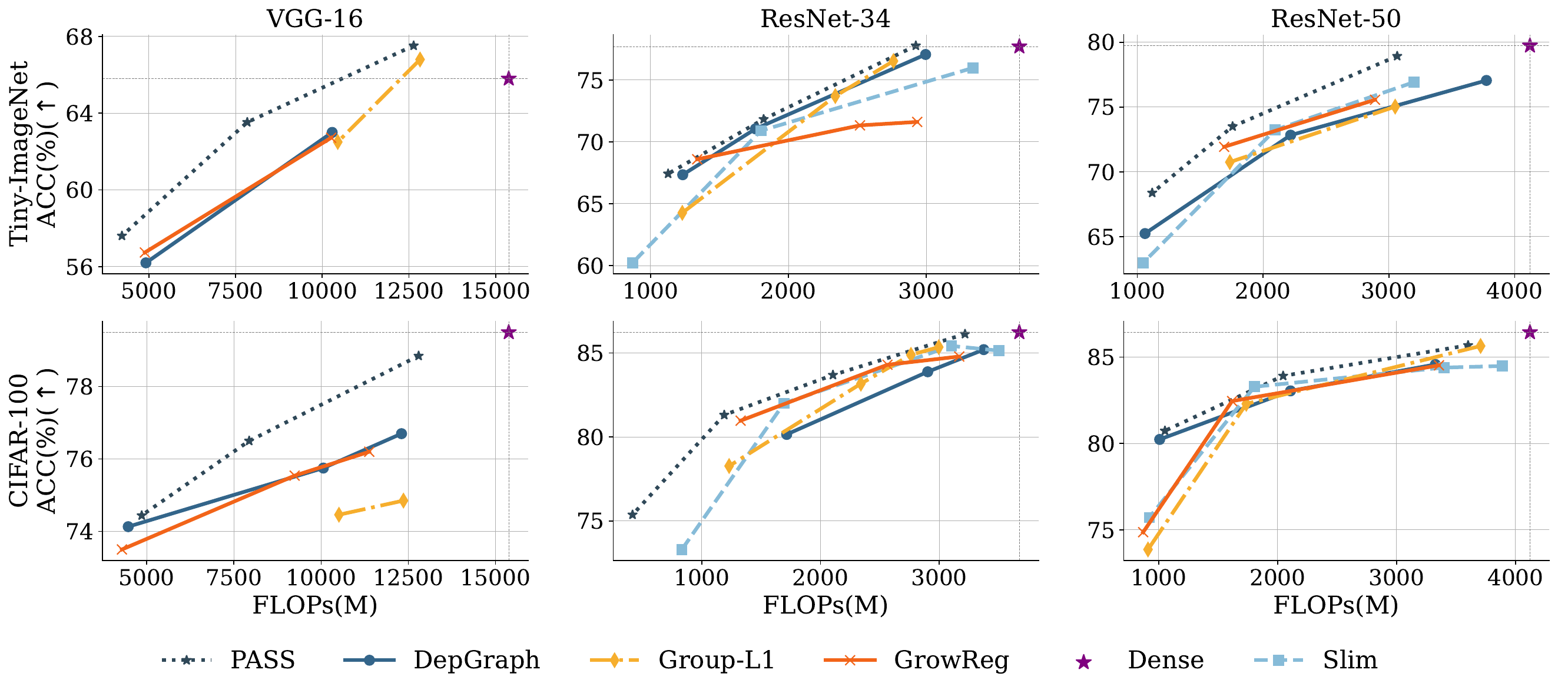}
\caption{Test accuracy of channel-pruned networks across various architectures based on CIFAR-$100$ and Tiny-ImageNet.}
\label{fig:architectures}
\vskip -0.15 in
\end{figure*}

\noindent\textbf{Superior Performance across Model Architectures.} We further evaluate the performance of \texttt{PASS} across multiple model architectures, namely VGG-$16$ without batch normalization~\footnote{The baseline Slim~\cite{liu2017learning} is not applicable to this architecture.}, ResNet-$34$, and ResNet-$50$ and compare it with the baselines. The results are shown in Figure~\ref{fig:architectures}. \underline{We observe} that our \texttt{PASS} achieves a competitive performance across all architectures, often achieving accuracy close to or even surpassing the dense models while being more computationally efficient.  For instance, To achieve an accuracy of  $75\%$ on Tiny-ImageNet using ResNet-$34$/ResNet-$50$ and $66\%$ accuracy using VGG-$16$, our \texttt{PASS} requires $0\%\sim12\%$ fewer FLOPs compared to the most efficient baseline. These observations suggest that \texttt{PASS} can effectively generalize across different architectures, maintaining a balance between computational efficiency and model performance.

\subsection{Experiments on ImageNet and Advanced Architectures}
To draw a solid conclusion, we further conduct extensive experiments on a large dataset ImageNet using advanced pre-trained models such as ResNeXt-50, Swin-T, and ViT-B/16.  The results are shown in Table~\ref{Tab:AdvancedModel}.  \underline{We observe} that our method \texttt{PASS} demonstrates a significant speed-up with minimal accuracy loss, as indicated by the $\Delta$ Acc., which is superior to existing methods like SSS~\cite{huang2018data}, GFP~\cite{liu2021group}, and DepGraph~\cite{fang2023depgraph}. the resulting empirical evidence robustly affirms the effectiveness of PASS across both advanced neural network architectures and large-scale datasets.

\subsection{Transferability of Learned Sparse Structure}
Inspired by studies suggesting the transferability of subnetworks between tasks~\cite{chen2020lottery}.  We investigate the transferability of \texttt{PASS} by posing two questions:($1$) \textit{Can the sparse channel masks, learned in one task, be effectively transferred to other tasks?} ($2$) \textit{Is the hypernetwork, once trained, applicable to other tasks?}   To answer Question ($1$), we test the accuracy of subnetworks found on Tiny-ImageNet when fine-tuning on CIFAR-$10/100$ and a pre-trained ResNet-18.  To answer Question ($2$), we measure the accuracy of the subnetwork finetuning on the target datasets, i.e., CIFAR-$10/100$. This subnetwork is obtained by applying hypernetworks, trained on Tiny-ImageNet, to the visual prompts of the respective target tasks. The results are reported in Table~\ref{tab:transfer}. \underline{We observe} that the channel mask and the hypernetwork, both learned by \texttt{PASS}, exhibit significant transferability on target datasets, highlighting their benefits across various subsequent tasks. More interestingly, the hypernetwork outperforms transferring the channel mask in most target tasks, providing two hints: \ding{182} Our learned hypernetworks can sufficiently capture the important topologies in downstream networks. Note that there is no parameter tuning for the hypernetworks and only with an adapted visual prompt. \ding{183} The visual prompt can effectively summarize the topological information from downstream neural networks, enabling superior sparsification.

\begin{table*}[htb]
\centering
\begin{adjustbox}{width=0.5\textwidth}
\begin{tabular}{lccccc}
\hline
\textbf{Arch.}   & \textbf{Method} & \textbf{Base} & \textbf{Pruned} & \textbf{$\Delta$ Acc.} & \textbf{FLOPs} \\ \hline
\multirow{5}{*}{ResNeXt-50} & ResNeXt-50 & 77.62 & -     & -    & 4.27 \\
           & SSS~\cite{huang2018data}   & 77.57 & 74.98 & -2.59 & 2.43 \\
           & GFP~\cite{liu2021group}   & 77.97 & 77.53 & -0.44 & 2.11 \\
           & DepGraph~\cite{fang2023depgraph}       & 77.62 & 76.48 & -1.14 & 2.09 \\ 
           & Ours (\texttt{PASS})      &  77.62& 77.21 &-0.41  &2.01 \\ \hline
\multirow{3}{*}{ViT-B/16}   & ViT-B/16   & 81.07 & -     & -    & 17.6 \\
           & DepGraph~\cite{fang2023depgraph}       & 81.07 & 79.17 & -1.90 & 10.4 \\ 
           & Ours(\texttt{PASS})      & 81.07 &79.77  &-1.30   & 10.7 \\ \hline
\multirow{4}{*}{Swin-T}      & Swin-T   & 81.4 & -     & -    & 4.49 \\
           &X-Pruner~\cite{yu2023x} &81.4& 80.7 & -0.7 & 3.2 \\
           &STEP~\cite{li2021differentiable}   &81.4&77.2&-4.2 & 3.5 \\
           & Ours(\texttt{PASS}) &81.4  &80.9 &-0.5 &3.4  \\ \hline  
\end{tabular}
\end{adjustbox}
\caption{Pruning results based on ImageNet and Advanced models.}
\vspace{-0.6em}
\label{Tab:AdvancedModel}
\end{table*}

\begin{table*}[tb]
\centering
\begin{adjustbox}{width=0.95\textwidth}
\begin{tabular}{l|cc|cc|cc}
\toprule
Channel Sparsity & \multicolumn{2}{c}{$10$\%} &\multicolumn{2}{c}{$30$\%} &\multicolumn{2}{c}{$50$\%}\\
\midrule
 & StanfordCars &CIFAR-100& StanfordCars &CIFAR-100& StanfordCars &CIFAR-100\\
\midrule
DepGraph&$75.79$&$81.60$&$69.26$ &$76.90$&$45.10$&$69.40$\\
Slim    &$58.10$&$80.27$&$43.00$ &$71.86$&$26.3$&$68.56$\\
Group-L1&$76.50$&$79.80$&$58.30$&$72.60$&$20.40$&$58.50$\\
Growreg    &$70.60$&$80.79$&$50.30$ &$72.27$&$41.80$&$65.80$\\
\hline
\gr Transfer Channel Mask &$83.50$&$82.45$&$79.70$&$80.83$&$76.60$&\textbf{$78.81$}\\
\gr Hypernetwork &\textbf{$84.31$}&\textbf{$82.49$}&\textbf{$79.88$}&\textbf{$80.98$}&\textbf{$76.80$}&$78.67$\\
\bottomrule
\end{tabular}
\end{adjustbox}
\caption{Transferability: Applying Channel Masks and Hypernetworks Learned from Tiny-ImageNet to CIFAR-$100$ and StanfordCars. The gray color denotes our method.}
\vspace{-0.3em}
\label{tab:transfer}
\end{table*}

\section{Ablations and Extra Invesitigations}


\noindent\textbf{Ablations on \texttt{PASS}}\label{ABonPASS}. To evaluate the effectiveness of \texttt{PASS}, we pose two interesting questions about the design of its components: ($1$) \textit{how do visual prompts and model weights contribute?} ($2$) \textit{is the recurrent mechanism crucial for mask finding?} To answer the above questions, we conduct a series of ablation studies utilizing a pre-trained ResNet-$18$ on CIFAR-$100$.  The extensive investigations contain ($1$) \textit{dropping either the visual prompt or model weights}; ($2$) \textit{destroy the recurrent nature in our hypernetwork}, such as using a Convolutional Neural Network (CNN) or a Multilayer Perceptron (MLP) to replace LSTM. The results are collected in Table~\ref{tab:ablations}. We observe that \ding{182} The exclusion of either the visual prompt or model weights leads to a pronounced drop in test accuracy (e.g., $83.45\%\rightarrow82.83\%$ and $82.66\%$ respectively at $90\%$ channel density), indicating the essential interplay role of both visual prompt and model weights in sparsification. \ding{183} If the recurrent nature in our design is destroyed, \textit{i.e.}, MLP or CNN methods variants, it suffers a performance decrement (\textit{e.g.}, $81.72\%\rightarrow$ $81.07\%$ and $81.09\%$ respectively at $70\%$ channel density). It implies a Markov property during the sparsification of two adjacent layers, which echoes the sparsity pathway findings in~\cite{wang2020picking}.
\begin{table*}[htb]
\centering
\begin{adjustbox}{width=0.75\textwidth}
\begin{tabular}{l|ccccc}
\toprule
\multicolumn{2}{c}{ Channel Sparsity $=$ } & $10$\% & $30$\% & $50$\% & $70$\% \\
\midrule
\multirow{3}{*}{Input Ablations} &LSTM+VP & $82.66$ & $81.20$ & $77.94$ & $72.01$ \\
&LSTM+Weights & $82.83$ & $81.13$ & $77.83$ & $72.45$ \\
\gr &LSTM+Weights+VP(Ours) & \textbf{$83.45$} & \textbf{$81.72$} & \textbf{$79.11$} & \textbf{$73.53$} \\
\hline
\multirow{3}{*}{Architecture Ablations} 
&ConVNet+VP & $83.21$ & $81.09$ & $78.15$ & $72.31$ \\
&MLP+VP+Weights & $83.23$ & $81.07$ & $77.84$ & $72.38$ \\
\gr&LSTM+Weights+VP(Ours) & \textbf{$83.45$} & \textbf{$81.72$} & \textbf{$79.11$} & \textbf{$73.53$} \\
\bottomrule
\end{tabular}
\end{adjustbox}
\caption{Ablations for \texttt{PASS} based on CIFAR-$100$ using a pre-trained ResNet-18.}
\vspace{-0.8em}
\label{tab:ablations}
\end{table*}

\begin{figure*}[htb]
\centering
\includegraphics[width=0.95\textwidth]{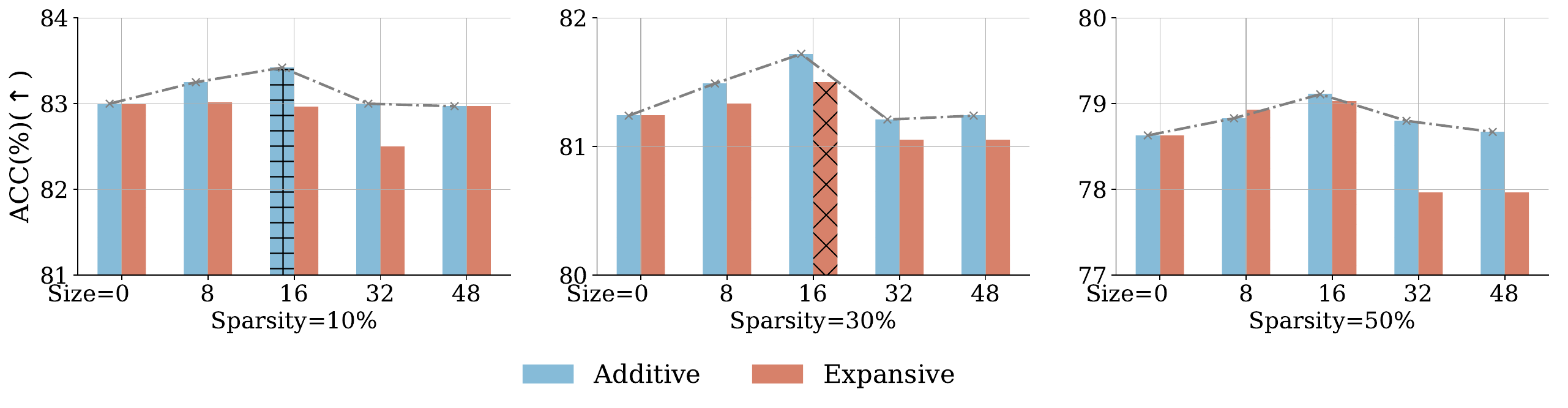}
\vskip -0.1 in
\caption{Ablation study on visual prompt strategies and their sizes. Experiments are conducted on CIFAR-100 and a pre-trained ResNet-18.}
\label{fig:VP_Strategy}
\end{figure*}

\noindent\textbf{Ablations on Visual Prompt}.
A visual prompt is a patch integrated with the input, as depicted in Figure~\ref{fig:framework}. Two prevalent methods for incorporating the visual prompt into the input have been identified in the literature~\cite{chen2023understanding,bahng2022exploring}:($1$) Adding to the input (abbreviated as~\textbf{``Additive visual prompt''}. ($2$)Expanding around the perimeter of the input, namely, the input is embedded into the central hollow section of the visual prompt (abbreviated as~\textbf{``Expansive visual prompt''}).  As discussed in section~\ref{ABonPASS}, visual prompt (VP) plays a key role in  \texttt{PASS}. Therefore, we pose such a question:\textit{How do the strategies and size of VP influence the performance of \texttt{PASS}?} To address this concern, we conduct experiments with ``Additive visual prompt'' and ``Expansive visual prompt'' respectively on CIFAR-$100$ using a pre-trained ResNet-$18$ under $10\%$, $30\%$ and $50\%$ channel sparsities, and we also show the performance of  \texttt{PASS} with varying the VP size from $0$ to $48$. The results are shown in Figure~\ref{fig:VP_Strategy}. We conclude that \ding{182} ``Additive visual prompt'' performs better than ``Expansive visual prompt'' across different sparsities. The disparity might be from the fact that ``Expansive visual prompt'' requires resizing the input to a smaller dimension, potentially leading to information loss, a problem that ``Additive visual prompt'' does not face. \ding{183}  The size of VP impacts the performance of \texttt{PASS}. We observe that test accuracy initially rises with the increase in VP size but starts to decline after reaching a peak at size $16$. A potential explanation for this decline is that the larger additive VP might overlap a significant portion of the input, leading to the loss of crucial information.  

\begin{figure*}[tb]
\centering
\includegraphics[width=0.95\textwidth]{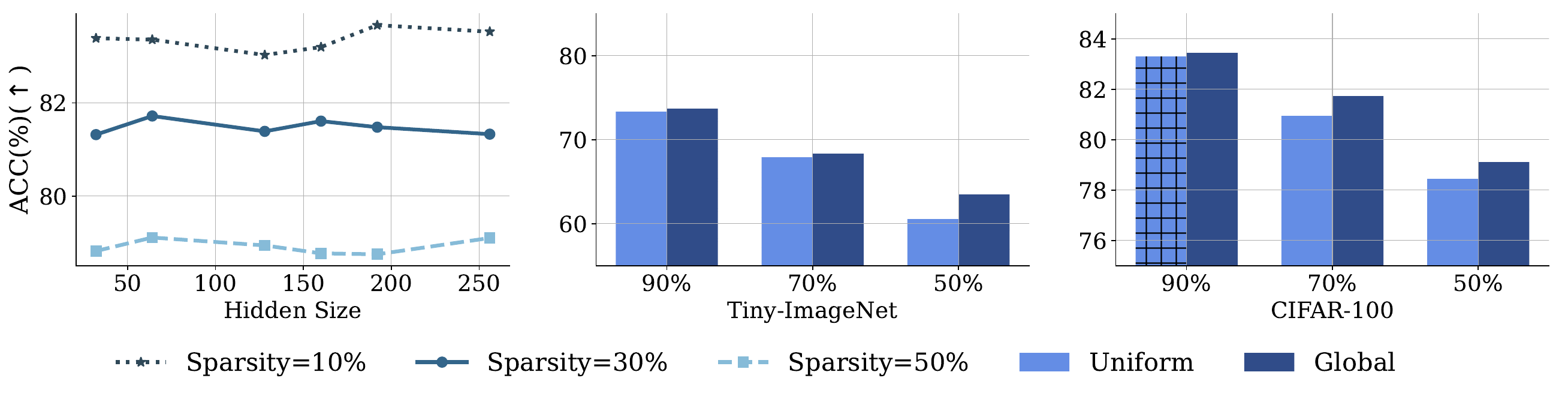}
\vskip -0.1 in
\caption{($1$)Ablation study of the hypernetwork's hidden size (Left Figure) using a pre-trained ResNet-$18$ on CIFAR-$100$. ($2$)Comparison between Global Pruning and Uniform Pruning strategies (Middle and Right Figures) using a pre-trained ResNet-18 on CIFAR-$100$ and Tiny-Imagenet.}
\label{fig:VP_Szie_GLobal_Uniform}
\vskip -0.15 in
\end{figure*}

\noindent\textbf{Impact of Hidden Size in HyperNetwork }
It is well-known that model size is an important factor impacting its performance, inducing the question \textit{how does the size of hypernetwork influence the performance of \textit{PASS}?}. To address this concern, we explore the impact of the hypernetwork hidden sizes on \texttt{PASS} by varying the hidden size of the proposed hypernetwork from $32$ to $256$ and evaluate its performance on CIFAR$100$ using a pre-trained ResNet-$18$ model under $10\%$, $30\%$ and $50\%$ channel sparsity respectively. The results are presented in the left figure of Figure~\ref{fig:VP_Szie_GLobal_Uniform}.  \underline{We observe} that the hidden size doesn't drastically affect the accuracy. While there are fluctuations, they are within a small range, suggesting that the hidden size is not a dominant factor in influencing the performance of \texttt{PASS}.

\noindent \textbf{Uniform Pruning VS Global Pruning.}
When converting the channel-wise importance scores into the channel masks, there are two prevalent strategies: ($1$) \textit{Uniform Pruning.}~\cite{ramanujan2020s,huang2022you} It prunes the channels of each layer with the lowest important scores by the same proportion.  ($2$) \textit{Global Pruning.}~\cite{huang2022you,fang2023depgraph} It prunes channels with the lowest important scores from all layers, leading to varied sparsity across layers. In this section, we evaluate the performance of global pruning and uniform pruning for \texttt{PASS} on CIFAR-$100$ using a pre-trained ResNet-$18$, with results presented in Figure~\ref{fig:VP_Szie_GLobal_Uniform}. \underline{We observe} that global pruning consistently yields higher test accuracy than uniform pruning, indicating its superior suitability for \texttt{PASS}, also reconfirming the importance of layer sparsity in sparsifying neural networks~\cite{liu2022the,huang2022you}.

\section{Conclusion}
In this paper, we delve deep into structural model pruning, with a particular focus on leveraging the potential of visual prompts for discerning channel importance in vision models. Our exploration highlights the key role of the input space and how judicious input editing can significantly influence the efficacy of structural pruning. We propose \texttt{PASS}, an innovative, end-to-end framework that harmoniously integrates visual prompts, providing a data-centric lens to channel pruning. Our recurrent mechanism adeptly addressed the intricate channel dependencies across layers, ensuring the derivation of high-quality structural sparsity.

Extensive evaluations across six datasets and four architectures underscore the prowess of \texttt{PASS}. The \texttt{PASS} framework excels not only in performance and computational efficiency but also demonstrates that its pruned models possess notable transferability. In essence, this research paves a new path for channel pruning, underscoring the importance of intertwining data-centric approaches with traditional model-centric methodologies. The fusion of these paradigms, as demonstrated by our findings, holds immense promise for the future of efficient neural network design.

\section*{Acknowledgements}

The authors acknowledge the use of resources provided by the Isambard-AI National AI Research Resource (AIRR) and the Dutch national e-infrastructure, supported by the SURF Cooperative (Project EINF-17091). Isambard-AI is operated by the University of Bristol and funded by the UK Government’s Department for Science, Innovation and Technology (DSIT) via UK Research and Innovation and the Science and Technology Facilities Council [ST/AIRR/I-A-I/1023]. Finally, we thank the anonymous reviewers for their insightful comments, which significantly improved the quality of this paper.

\bibliography{cpal_2026}

@article{bahng2022exploring,
  title={Exploring visual prompts for adapting large-scale models},
  author={Bahng, Hyojin and Jahanian, Ali and Sankaranarayanan, Swami and Isola, Phillip},
  journal={arXiv preprint arXiv:2203.17274},
  volume={1},
  number={3},
  pages={4},
  year={2022}
}

@inproceedings{he2018amc,
  title={Amc: Automl for model compression and acceleration on mobile devices},
  author={He, Yihui and Lin, Ji and Liu, Zhijian and Wang, Hanrui and Li, Li-Jia and Han, Song},
  booktitle={Proceedings of the European conference on computer vision (ECCV)},
  pages={784--800},
  year={2018}
}

@inproceedings{dehghani2023scaling,
  title={Scaling vision transformers to 22 billion parameters},
  author={Dehghani, Mostafa and Djolonga, Josip and Mustafa, Basil and Padlewski, Piotr and Heek, Jonathan and Gilmer, Justin and Steiner, Andreas Peter and Caron, Mathilde and Geirhos, Robert and Alabdulmohsin, Ibrahim and others},
  booktitle={International Conference on Machine Learning},
  pages={7480--7512},
  year={2023},
  organization={PMLR}
}

@article{pham2022understanding,
  title={Understanding Pruning at Initialization: An Effective Node-Path Balancing Perspective},
  author={Pham, Hoang and Ta, Anh and Liu, Shiwei and Le, Dung D and Tran-Thanh, Long},
  year={2022}
}

@inproceedings{chen2023understanding,
  title={Understanding and improving visual prompting: A label-mapping perspective},
  author={Chen, Aochuan and Yao, Yuguang and Chen, Pin-Yu and Zhang, Yihua and Liu, Sijia},
  booktitle={Proceedings of the IEEE/CVF Conference on Computer Vision and Pattern Recognition},
  pages={19133--19143},
  year={2023}
}

@inproceedings{xiao2023smoothquant,
  title={Smoothquant: Accurate and efficient post-training quantization for large language models},
  author={Xiao, Guangxuan and Lin, Ji and Seznec, Mickael and Wu, Hao and Demouth, Julien and Han, Song},
  booktitle={International Conference on Machine Learning},
  pages={38087--38099},
  year={2023},
  organization={PMLR}
}

@article{dettmers2022llm,
  title={Llm. int8 (): 8-bit matrix multiplication for transformers at scale},
  author={Dettmers, Tim and Lewis, Mike and Belkada, Younes and Zettlemoyer, Luke},
  journal={arXiv preprint arXiv:2208.07339},
  year={2022}
}

@article{wang2020picking,
  title={Picking winning tickets before training by preserving gradient flow},
  author={Wang, Chaoqi and Zhang, Guodong and Grosse, Roger},
  journal={arXiv preprint arXiv:2002.07376},
  year={2020}
}

@article{liu2023pre,
  title={Pre-train, prompt, and predict: A systematic survey of prompting methods in natural language processing},
  author={Liu, Pengfei and Yuan, Weizhe and Fu, Jinlan and Jiang, Zhengbao and Hayashi, Hiroaki and Neubig, Graham},
  journal={ACM Computing Surveys},
  volume={55},
  number={9},
  pages={1--35},
  year={2023},
  publisher={ACM New York, NY}
}

@inproceedings{liu2023explicit,
  title={Explicit visual prompting for low-level structure segmentations},
  author={Liu, Weihuang and Shen, Xi and Pun, Chi-Man and Cun, Xiaodong},
  booktitle={Proceedings of the IEEE/CVF Conference on Computer Vision and Pattern Recognition},
  pages={19434--19445},
  year={2023}
}

@article{zang2022unified,
  title={Unified vision and language prompt learning},
  author={Zang, Yuhang and Li, Wei and Zhou, Kaiyang and Huang, Chen and Loy, Chen Change},
  journal={arXiv preprint arXiv:2210.07225},
  year={2022}
}

@article{brown2020language,
  title={Language models are few-shot learners},
  author={Brown, Tom and Mann, Benjamin and Ryder, Nick and Subbiah, Melanie and Kaplan, Jared D and Dhariwal, Prafulla and Neelakantan, Arvind and Shyam, Pranav and Sastry, Girish and Askell, Amanda and others},
  journal={Advances in neural information processing systems},
  volume={33},
  pages={1877--1901},
  year={2020}
}

@article{shin2020autoprompt,
  title={Autoprompt: Eliciting knowledge from language models with automatically generated prompts},
  author={Shin, Taylor and Razeghi, Yasaman and Logan IV, Robert L and Wallace, Eric and Singh, Sameer},
  journal={arXiv preprint arXiv:2010.15980},
  year={2020}
}

@article{li2021prefix,
  title={Prefix-tuning: Optimizing continuous prompts for generation},
  author={Li, Xiang Lisa and Liang, Percy},
  journal={arXiv preprint arXiv:2101.00190},
  year={2021}
}

@article{molchanov2016pruning,
  title={Pruning convolutional neural networks for resource efficient inference},
  author={Molchanov, Pavlo and Tyree, Stephen and Karras, Tero and Aila, Timo and Kautz, Jan},
  journal={arXiv preprint arXiv:1611.06440},
  year={2016}
}

@article{chen2020lottery,
  title={The lottery ticket hypothesis for pre-trained bert networks},
  author={Chen, Tianlong and Frankle, Jonathan and Chang, Shiyu and Liu, Sijia and Zhang, Yang and Wang, Zhangyang and Carbin, Michael},
  journal={Advances in neural information processing systems},
  volume={33},
  pages={15834--15846},
  year={2020}
}

@article{xu2023compress,
  title={Compress, Then Prompt: Improving Accuracy-Efficiency Trade-off of LLM Inference with Transferable Prompt},
  author={Xu, Zhaozhuo and Liu, Zirui and Chen, Beidi and Tang, Yuxin and Wang, Jue and Zhou, Kaixiong and Hu, Xia and Shrivastava, Anshumali},
  journal={arXiv preprint arXiv:2305.11186},
  year={2023}
}

@inproceedings{liu2017learning,
  title={Learning efficient convolutional networks through network slimming},
  author={Liu, Zhuang and Li, Jianguo and Shen, Zhiqiang and Huang, Gao and Yan, Shoumeng and Zhang, Changshui},
  booktitle={Proceedings of the IEEE international conference on computer vision},
  pages={2736--2744},
  year={2017}
}

@inproceedings{jia2022visual,
  title={Visual prompt tuning},
  author={Jia, Menglin and Tang, Luming and Chen, Bor-Chun and Cardie, Claire and Belongie, Serge and Hariharan, Bharath and Lim, Ser-Nam},
  booktitle={European Conference on Computer Vision},
  pages={709--727},
  year={2022},
  organization={Springer}
}

@article{zha2023data,
  title={Data-centric artificial intelligence: A survey},
  author={Zha, Daochen and Bhat, Zaid Pervaiz and Lai, Kwei-Herng and Yang, Fan and Jiang, Zhimeng and Zhong, Shaochen and Hu, Xia},
  journal={arXiv preprint arXiv:2303.10158},
  year={2023}
}

@inproceedings{fang2023depgraph,
  title={Depgraph: Towards any structural pruning},
  author={Fang, Gongfan and Ma, Xinyin and Song, Mingli and Mi, Michael Bi and Wang, Xinchao},
  booktitle={Proceedings of the IEEE/CVF Conference on Computer Vision and Pattern Recognition},
  pages={16091--16101},
  year={2023}
}

@inproceedings{
li2017pruning,
title={Pruning Filters for Efficient ConvNets},
author={Hao Li and Asim Kadav and Igor Durdanovic and Hanan Samet and Hans Peter Graf},
booktitle={International Conference on Learning Representations},
year={2017},
url={https://openreview.net/forum?id=rJqFGTslg}
}

@inproceedings{
wang2021neural,
title={Neural Pruning via Growing Regularization},
author={Huan Wang and Can Qin and Yulun Zhang and Yun Fu},
booktitle={International Conference on Learning Representations},
year={2021},
url={https://openreview.net/forum?id=o966_Is_nPA}
}

@article{lin2020channel,
  title={Channel pruning via automatic structure search},
  author={Lin, Mingbao and Ji, Rongrong and Zhang, Yuxin and Zhang, Baochang and Wu, Yongjian and Tian, Yonghong},
  journal={arXiv preprint arXiv:2001.08565},
  year={2020}
}

@article{bengio2013estimating,
  title={Estimating or propagating gradients through stochastic neurons for conditional computation},
  author={Bengio, Yoshua and L{\'e}onard, Nicholas and Courville, Aaron},
  journal={arXiv preprint arXiv:1308.3432},
  year={2013}
}

@article{chiang2023vicuna,
  title={Vicuna: An open-source chatbot impressing gpt-4 with 90\%* chatgpt quality},
  author={Chiang, Wei-Lin and Li, Zhuohan and Lin, Zi and Sheng, Ying and Wu, Zhanghao and Zhang, Hao and Zheng, Lianmin and Zhuang, Siyuan and Zhuang, Yonghao and Gonzalez, Joseph E and others},
  journal={Arxiv preprint},
  year={2023}
}

@article{bai2023qwen,
  title={Qwen-VL: A Frontier Large Vision-Language Model with Versatile Abilities},
  author={Bai, Jinze and Bai, Shuai and Yang, Shusheng and Wang, Shijie and Tan, Sinan and Wang, Peng and Lin, Junyang and Zhou, Chang and Zhou, Jingren},
  journal={arXiv preprint arXiv:2308.12966},
  year={2023}
}

@article{huang2023large,
  title={Are Large Kernels Better Teachers than Transformers for ConvNets?},
  author={Huang, Tianjin and Yin, Lu and Zhang, Zhenyu and Shen, Li and Fang, Meng and Pechenizkiy, Mykola and Wang, Zhangyang and Liu, Shiwei},
  journal={ICML},
  year={2023}
}

@article{sun2019patient,
  title={Patient knowledge distillation for bert model compression},
  author={Sun, Siqi and Cheng, Yu and Gan, Zhe and Liu, Jingjing},
  journal={arXiv preprint arXiv:1908.09355},
  year={2019}
}

@article{chen2022improving,
  title={Improving In-Context Few-Shot Learning via Self-Supervised Training},
  author={Chen, Mingda and Du, Jingfei and Pasunuru, Ramakanth and Mihaylov, Todor and Iyer, Srini and Stoyanov, Veselin and Kozareva, Zornitsa},
  journal={arXiv preprint arXiv:2205.01703},
  year={2022}
}

@article{razdaibiedina2023progressive,
  title={Progressive prompts: Continual learning for language models},
  author={Razdaibiedina, Anastasia and Mao, Yuning and Hou, Rui and Khabsa, Madian and Lewis, Mike and Almahairi, Amjad},
  journal={arXiv preprint arXiv:2301.12314},
  year={2023}
}

@inproceedings{chen2022knowprompt,
  title={Knowprompt: Knowledge-aware prompt-tuning with synergistic optimization for relation extraction},
  author={Chen, Xiang and Zhang, Ningyu and Xie, Xin and Deng, Shumin and Yao, Yunzhi and Tan, Chuanqi and Huang, Fei and Si, Luo and Chen, Huajun},
  booktitle={Proceedings of the ACM Web conference 2022},
  pages={2778--2788},
  year={2022}
}

@inproceedings{
liu2022the,
title={The Unreasonable Effectiveness of Random Pruning: Return of the Most Naive Baseline for Sparse Training},
author={Shiwei Liu and Tianlong Chen and Xiaohan Chen and Li Shen and Decebal Constantin Mocanu and Zhangyang Wang and Mykola Pechenizkiy},
booktitle={International Conference on Learning Representations},
year={2022},
url={https://openreview.net/forum?id=VBZJ_3tz-t}
}

@article{huang2022you,
  title={You Can Have Better Graph Neural Networks by Not Training Weights at All: Finding Untrained GNNs Tickets},
  author={Huang, Tianjin and Chen, Tianlong and Fang, Meng and Menkovski, Vlado and Zhao, Jiaxu and Yin, Lu and Pei, Yulong and Mocanu, Decebal Constantin and Wang, Zhangyang and Pechenizkiy, Mykola and others},
  journal={arXiv preprint arXiv:2211.15335},
  year={2022}
}

@article{he2017channel,
  title={Channel pruning for accelerating very deep neural networks},
  author={He, Yihui and Zhang, Xiangyu and Sun, Jian},
  journal={Proceedings of the IEEE International Conference on Computer Vision},
  pages={1389--1397},
  year={2017}
}

@article{hou2020feature,
  title={A Feature-map Discriminant Perspective for Pruning Deep Neural Networks},
  author={Hou, Zejiang and Kung, Sun-Yuan},
  journal={arXiv preprint arXiv:2005.13796},
  year={2020}
}

@article{ye2020good,
  title={Good subnetworks provably exist: Pruning via greedy forward selection},
  author={Ye, Mao and Gong, Chengyue and Nie, Lizhen and Zhou, Denny and Klivans, Adam and Liu, Qiang},
  journal={Proceedings of International Conference on Machine Learning},
  pages={10820--10830},
  year={2020},
}

@article{luo2020neural,
  title={Neural Network Pruning with Residual-Connections and Limited-Data},
  author={Luo, Jian-Hao and Wu, Jianxin},
  journal={Proceedings of the IEEE/CVF Conference on Computer Vision and Pattern Recognition},
  pages={1458--1467},
  year={2020}
}

@article{li2016pruning,
  title={Pruning filters for efficient convnets},
  author={Li, Hao and Kadav, Asim and Durdanovic, Igor and Samet, Hanan and Graf, Hans Peter},
  journal={arXiv preprint arXiv:1608.08710},
  year={2016}
}

@article{zhang2018graph,
  title={Graph hypernetworks for neural architecture search},
  author={Zhang, Chris and Ren, Mengye and Urtasun, Raquel},
  journal={arXiv preprint arXiv:1810.05749},
  year={2018}
}

@article{chauhan2023dynamic,
  title={Dynamic inter-treatment information sharing for heterogeneous treatment effects estimation},
  author={Chauhan, Vinod Kumar and Zhou, Jiandong and Molaei, Soheila and Ghosheh, Ghadeer and Clifton, David A},
  journal={arXiv preprint arXiv:2305.15984},
  year={2023}
}

@inproceedings{zhao2020meta,
  title={Meta-learning via hypernetworks},
  author={Zhao, Dominic and Kobayashi, Seijin and Sacramento, Jo{\~a}o and von Oswald, Johannes},
  booktitle={4th Workshop on Meta-Learning at NeurIPS 2020 (MetaLearn 2020)},
  year={2020},
  organization={NeurIPS}
}

@inproceedings{ramanujan2020s,
  title={What's hidden in a randomly weighted neural network?},
  author={Ramanujan, Vivek and Wortsman, Mitchell and Kembhavi, Aniruddha and Farhadi, Ali and Rastegari, Mohammad},
  booktitle={Proceedings of the IEEE/CVF conference on computer vision and pattern recognition},
  pages={11893--11902},
  year={2020}
}

@article{han2021dynamic,
  title={Dynamic neural networks: A survey},
  author={Han, Yizeng and Huang, Gao and Song, Shiji and Yang, Le and Wang, Honghui and Wang, Yulin},
  journal={IEEE Transactions on Pattern Analysis and Machine Intelligence},
  volume={44},
  number={11},
  pages={7436--7456},
  year={2021},
  publisher={IEEE}
}

@inproceedings{huang2018data,
  title={Data-driven sparse structure selection for deep neural networks},
  author={Huang, Zehao and Wang, Naiyan},
  booktitle={Proceedings of the European conference on computer vision (ECCV)},
  pages={304--320},
  year={2018}
}

@inproceedings{liu2021group,
  title={Group fisher pruning for practical network compression},
  author={Liu, Liyang and Zhang, Shilong and Kuang, Zhanghui and Zhou, Aojun and Xue, Jing-Hao and Wang, Xinjiang and Chen, Yimin and Yang, Wenming and Liao, Qingmin and Zhang, Wayne},
  booktitle={International Conference on Machine Learning},
  pages={7021--7032},
  year={2021},
  organization={PMLR}
}

@inproceedings{yu2023x,
  title={X-Pruner: eXplainable Pruning for Vision Transformers},
  author={Yu, Lu and Xiang, Wei},
  booktitle={Proceedings of the IEEE/CVF Conference on Computer Vision and Pattern Recognition},
  pages={24355--24363},
  year={2023}
}

@article{li2021differentiable,
  title={Differentiable subset pruning of transformer heads},
  author={Li, Jiaoda and Cotterell, Ryan and Sachan, Mrinmaya},
  journal={Transactions of the Association for Computational Linguistics},
  volume={9},
  pages={1442--1459},
  year={2021},
  publisher={MIT Press One Rogers Street, Cambridge, MA 02142-1209, USA journals-info~…}
}


\appendix
\newpage
\section{Parameters of Hypernetworks}
In this study, the hidden size of the hypernetwork is configured to $64$. A detailed breakdown of the number of parameters for the hypernetworks utilized in this research is provided in Table~\ref{tab:Nparameters}. It is noteworthy that the parameter count for the hypernetworks is significantly lower compared to that of the pretrained models. For instance, in the case of ResNet-$18$, the hypernetwork parameters constitute only $2.8\%$ of the total parameters of the pre-trained ResNet-$18$.

\begin{table}[!htb]
\centering
\caption{The number of parameters for our Hypernetworks.} 
\label{tab:Nparameters}
\resizebox{0.8\textwidth}{!}{%
\begin{tabular}{@{}l|cccccc@{}}
\toprule
 & \multicolumn{1}{c}{ResNet-18 (11M)} & \multicolumn{1}{c}{ResNet-34 (21M)} & \multicolumn{1}{c}{ResNet-50 (25M)} & \multicolumn{1}{c}{VGG-16 (138M)} \\  \midrule
    \#Parameters-HyperNetwork&0.31M (2.8\%)&0.56M (2.6\%) &1.5M (6\%)&0.34M (0.2\%) \\
\bottomrule
\end{tabular}}
\end{table}

\section{Implementation Details}\label{ImplementDetails}
Table~\ref{tab:implementations} summarizes the hyper-parameters for \texttt{PASS} used in all our experiments.   

\begin{table}[!htb]
\centering
\vspace{-4mm}
\caption{Implementation details on each dataset.}
\label{tab:implementations}
\begin{adjustbox}{width=1\textwidth}
\begin{threeparttable}
\begin{tabular}{l|cccccc}
\toprule
\multirow{1}{*}{Settings} & \multicolumn{1}{c}{Tny-ImageNet} & \multicolumn{1}{c}{CIFAR-10} & \multicolumn{1}{c}{CIFAF-10} & \multicolumn{1}{c}{DTD}& \multicolumn{1}{c}{StanfordCars}& \multicolumn{1}{c}{Food101}\\  \midrule
\multicolumn{7}{c}{Stage 1: Learning to Prune}\\
\hline
Batch Size &\multicolumn{6}{c}{$128$} \\ \midrule
Weight Decay - VP & $0$ &$0$&$0$&$0$&$0$&$0$\\ 
Learning Rate - VP& $1e-2$&$1e-2$&$1e-2$&$1e-2$&$1e-2$&$1e-2$\\
Optimizer - VP &\multicolumn{6}{c}{SGD optimizer}\\
LR-Decay-Scheduler - VP&\multicolumn{6}{c}{cosine}\\ \midrule
Weight Decay - HyperNetwork & $1e-2$  &$1e-2$&$1e-2$&$1e-2$&$1e-2$&$1e-2$\\ 
Learning Rate - HyperNetwork& $1e-3$&$1e-3$&$1e-3$&$1e-3$&$1e-3$&$1e-3$\\
Optimizer - HyperNetwork& \multicolumn{6}{c}{AdamW optimizer}\\
LR-Decay-Scheduler - HyperNetwork&\multicolumn{6}{c}{cosine}\\
Total epochs & \multicolumn{6}{c}{$50$}\\
\hline
\hline
\multicolumn{7}{c}{Stage 2: Fine-tune}\\
\midrule
Batch Size &\multicolumn{6}{c}{$128$} \\ \midrule
Weight Decay - VP & $0$ &$0$&$0$&$0$&$0$&$0$\\ 
Learning Rate - VP& $1e-3$&$1e-2$&$1e-2$&$1e-2$&$1e-2$&$1e-2$\\
Optimizer - VP &\multicolumn{6}{c}{SGD optimizer}\\
LR-Decay-Scheduler - VP&\multicolumn{6}{c}{cosine}\\ \midrule
Weight Decay - Pruned Network & $5e-4$  &$3e-4$&$5e-4$&$5e-4$&$5e-4$&$5e-4$\\ 
Learning Rate - Pruned Network& $1e-3$&$1e-2$&$1e-2$&$1e-2$&$1e-2$&$1e-2$\\
Optimizer - Pruned Network& \multicolumn{6}{c}{SGD optimizer}\\
LR-Decay-Scheduler - Pruned Network&multistep-$\{6,8\}$&cosine&cosine&cosine&cosine&cosine\\
Total epochs & $10$ &$50$&$50$&$50$&$50$&$50$\\
\bottomrule
\end{tabular}
\end{threeparttable}
\end{adjustbox}
\vspace{-4mm}
\end{table}

\section{Learned Channel Sparsity}\label{Learned Density}
We present the channel sparsity learned by \texttt{PASS} on CIFAR-$100$ and Tiny-ImageNet using a pre-trained ResNet-$18$ in Table~\ref{tab:res18sparsity}. Our observations indicate that channel sparsity is generally higher in the top layers and lower in the bottom layers of the network.

\begin{table}[!htb]
\vspace{-6mm}
\centering
\caption{Layer-wise sparsity of the pre-trained ResNet-$18$ on CIFAR-$100$ and Tiny-ImageNet as learned by \texttt{PASS} at $30\%, 50\%$ sparsity levels.}
\label{tab:res18sparsity}
\resizebox{\textwidth}{!}{%
\begin{tabular}{@{}l|c|cc|cc@{}}
\toprule
\multirow{2}{*}{Layer} & \multirow{2}{*}{Fully Dense \#Channels} & \multicolumn{2}{c}{CIFAR-100} & \multicolumn{2}{c}{Tiny-ImageNet} \\
& & $30\%$ Sparsity & $50\%$ Sparsity & $30\%$ Sparsity & $50\%$ Sparsity \\
\midrule
Layer 1 - conv1 & $64$ & $9.4\%$ & $29.7\%$ & $20.3\%$ & $37.5\%$ \\
Layer 2 - layer1.0.conv1 & $64$ & $17.2\%$ & $43.8\%$ &$28.1\%$ & $56.2\%$ \\
Layer 3 - layer1.0.conv2 & $64$ & $29.7\%$ & $29.7\%$ & $20.3\%$ & $39.0\%$ \\
Layer 4 - layer1.1.conv1 & $64$ & $15.7\%$ & $46.9\%$ & $50\%$ & $62.5\%$ \\
Layer 5 - layer1.1.conv2 & $64$ & $22.7\%$ & $26.6\%$ & $17.1\%$ & $32.8\%$ \\
Layer 6 - layer2.0.conv1 & $128$ & $19.6\%$ & $46.9\%$ & $42.9\%$ & $57.0\%$ \\
Layer 7 - layer2.0.conv2 & $128$ & $19.6\%$ & $45.4\%$ & $14.0\%$ & $34.3\%$ \\
Layer 8 - layer2.0.downsample.0 & $128$ & $19.6\%$ & $45.4\%$ & $14.0\%$ & $34.3\%$ \\
Layer 9 - layer2.1.conv1 & $128$ & $19.6\%$ & $44.6\%$ & $34.3\%$ & $55.5\%$ \\
Layer 10 - layer2.1.conv2 & $128$ & $7.1\%$ & $28.9\%$ & $16.4\%$ & $34.3\%$ \\
Layer 11 - layer3.0.conv1 & $256$ & $29\%$ & $50\%$ & $42.1\%$ & $58.9\%$ \\
Layer 12 - layer3.0.conv2 & $256$ & $15.3\%$ & $50\%$ & $11.7\%$ &$28.5\%$ \\
Layer 13 - layer3.0.downsample.0 & $256$ & $15.3\%$ & $50\%$ & $11.7\%$ & $28.5\%$ \\
Layer 14 - layer3.1.conv1 & $256$ & $27.4\%$ & $43.8\%$ & $33.2\%$ & $52.3\%$ \\
Layer 15 - layer3.1.conv2 & $256$ & $11\%$ & $26.2\%$ & $10.1\%$ & $23.8\%$ \\
Layer 16 - layer4.0.conv1 & $512$ & $29.3\%$ & $50\%$ & $33.3\%$ & $54.4\%$ \\
Layer 17 - layer4.0.conv2 & $512$ & $41.8\%$ & $50\%$ & $36.1\%$ & $58.9\%$ \\
Layer 18 - layer4.0.downsample.0 & $512$ & $41.8\%$ & $50\%$ & $36.1\%$ & $58.9\%$ \\
Layer 19 - layer4.1.conv1 & $512$ & $44.6\%$ & $48.3\%$ & $39.2\%$ & $65.8\%$ \\
Layer 20 - layer4.1.conv2 & $512$ & $42.6\%$ & $49.9\%$ & $27.1\%$ & $46.2\%$ \\
Layer 21 - Linear & $512$ & $0\%$ & $0\%$ & $0\%$ & $0\%$ \\
\bottomrule
\end{tabular}}
\vspace{-4mm}
\end{table}

\section{Complexity Analysis of the Hypernetwork}

In this section, we provide a comprehensive analysis about the complexity of the Hypernetwork. (1) Regarding the impact on time complexity, our recurrent hyper-network is designed for efficiency. The channel masks are pre-calculated, eliminating the need for real-time generation during both the inference and subnetwork fine-tuning phases. Therefore, the recurrent hyper-network does not introduce any extra time complexity during the inference and the fintune-tuning phase. The additional computing time is limited to the phase of channel mask identification. (2) Moreover, the hyper-network itself is designed to be lightweight. The number of parameters it contributes to the overall model is minimal, thus ensuring that any additional complexity during the mask-finding phase is negligible. This claim is substantiated by empirical observations: the hyper-network accounts for only about $0.2$\% to $6$\% of the total model parameters across various architectures such as ResNet-18/34 and VGG-16, as illustrated in Table~\ref{tab:Nparameters}. (3) Additionally, we assessed the training time per epoch with and without the hyper-network during the channel mask identification phase. Our findings in Table~\ref{tab:time} indicate that the inclusion of the LSTM network has a marginal effect on these durations, further affirming the efficiency of our approach.

\begin{table}[!htb]
\vspace{-2mm}
\centering
\caption{Per-Epoch Training Time (s) With/Without Hypernetworks in Channel Mask Identification (1× A100)}
\label{tab:time}
\resizebox{0.8\textwidth}{!}{%
\begin{tabular}{@{}l|ccccc@{}}
\toprule
 & \multicolumn{1}{c}{ResNet-18 (11M)} & \multicolumn{1}{c}{ResNet-34 (21M)} & \multicolumn{1}{c}{ResNet-50 (25M)}  \\  \midrule
    w/o HyperNetwork&70.05&73.95 &95.65 \\
    w/ HyperNetwork&72.2&76.95 &111.6\\
\bottomrule
\end{tabular}}
\end{table}

\section{Difference between our proposed PASS and dynamic neural network}
There are two fundamental differences between our proposed PASS and dynamic neural network. (1) \textbf{The hyper-network in our proposed PASS is not `dynamic'.} Dynamic neural networks, as categorized in the literature, are networks capable of adapting their structures or parameters conditioned in a sample-dependent manner, as outlined in~\cite{han2021dynamic}. In contrast, the hyper-network within our PASS framework does not exhibit this `dynamic' nature. It is designed to be dependent on a visual prompt (task-specific), as opposed to dynamically adjusting to input samples.  This hyper-network's role is confined to the channel mask identification phase and is not employed during the inference phase. Therefore, it is fundamentally different from dynamic neural networks.  (2) \textbf{Their goals are different.} The fundamental goal of the hyper-network in PASS is distinct from that of dynamic neural networks. While the latter focuses on adapting their architecture or parameters based on input samples, our hyper-network is specifically engineered for the integration of visual prompts with the statics of model weights.

\end{document}